\documentclass{article}
\usepackage{spconf,amsmath,graphicx}
\usepackage{bbding}

\usepackage{subfigure}

\title{Learning Enhanced Resolution-wise features for \\Human Pose Estimation}
\name{Kun. Zhang$^{1*}$, Peng He$^{2}$\sthanks{$^*$ These two authors contribute equally to this paper.}, Ping Yao$^{1}$\sthanks{$^\dagger$  Corresponding author.},\thanks{This work is supported by Strategic Priority Research Program of the Chinese Academy of Sciences (Grant No.XDA19020400), Equipment Pre-Research Fund (Grant No.61403120405, Grant No.6141B07090131), and Spaceborne Equipment Pre-Research Project (Grant No. 305030704).} 
Ge Chen$^{1}$, Rui Wu$^{4}$, Min Du$^{4}$, Huimin Li$^{3}$, Li Fu$^{1}$, Tianyao Zheng$^{1}$}
\address{$^{1}$Institute of Computing Technology, Chinese Academy of Sciences\\
$^{2}$School of Computer Science and Technology, University of Chinese Academy of Sciences\\
$^{3}$School of Automation, Beijing Institute of Technology\\
$^{4}$Horizon Robotics}
\begin{document}
%
\maketitle
\begin{abstract}
Recently, multi-resolution networks (such as Hourglass, CPN, HRNet, etc.) have achieved significant performance on pose estimation by combining feature maps of various resolutions. In this paper, we propose a Resolution-wise Attention Module (RAM) and Gradual Pyramid Refinement (GPR), to learn enhanced resolution-wise feature maps for precise pose estimation. Specifically, RAM learns a group of weights to represent the different importance of feature maps across resolutions, and the GPR gradually merges every two feature maps from low to high resolutions to regress final human keypoint heatmaps. With the enhanced resolution-wise features learnt by CNN, we obtain more accurate human keypoint locations. \\
The efficacies of our proposed methods are demonstrated on MS-COCO dataset, achieving state-of-the-art performance with average precision of 77.7 on COCO val2017 set and 77.0 on test-dev2017 set without using extra human keypoint training dataset.
\end{abstract}
\begin{keywords}Human Pose Estimation, Multi-Resolution Network, Attention Mechanism\end{keywords}
\section{Introduction}
\label{sec:intro}

2D pose estimation refers to the task of providing accurate locations for human keypoints (such as heads, knees, ankles, shoulders, wrists, etc.) from two-dimensional digital images, which remains important but challenging problem in the field of image processing. This paper focus on single-person pose estimation, which stays as an active research topic for decades. Pose estimation also serves as a basic technology to deal with several practical applications, such as human computer interaction \cite{stgcn}, human action recognition \cite{humanfly},  and pose tracking \cite{simplebaseline}. In recent years, the progress on human pose estimation has been benefited by representation power of neural networks \cite{resnet,CPM}.

Accurate keypoint estimation networks need not only low-level feature maps from high resolutions for precise pixel-level keypoint heatmap prediction, but also high-level features with large receptive fields for invisible human keypoint inference. State-of-the-art approaches for human pose estimation tend to fuse multi-resolution features to satisfy these two requirements. For instance, Hourglass-based pose estimation networks \cite{hourglasse1,hourglass} capture and consolidate feature maps across various scales with repeated symmetric network structures; Pyramid-based networks \cite{cpn1,seu2019} integrate information from different ResNet stages via GlobalNet and RefineNet; HRNet \cite{hrv1} keeps high-resolution representations through the whole network architecture, and fuses multi-scale features in stages to enrich feature representations. These networks have achieved significant performance on the task of human pose estimation by fusing feature maps from different resolutions.

In order to enhance features from different resolutions more efficiently, we propose two new modules: Resolution-wise Attention Module and Gradual Pyramid Refinement. To be specific, each Resolution-wise Attention Module (RAM) learns a group of weights to represent importance of features from each resolution, and then sums interpolated feature maps from different resolutions according to the learned weights for resolution-wise information exchange. The Gradual Pyramid Refinement (GPR) module recursively upsamples and merges feature maps of low to high resolution to regress human keypoint heatmaps. 

\begin{figure*}
\centerline{\includegraphics[width=17.5cm]{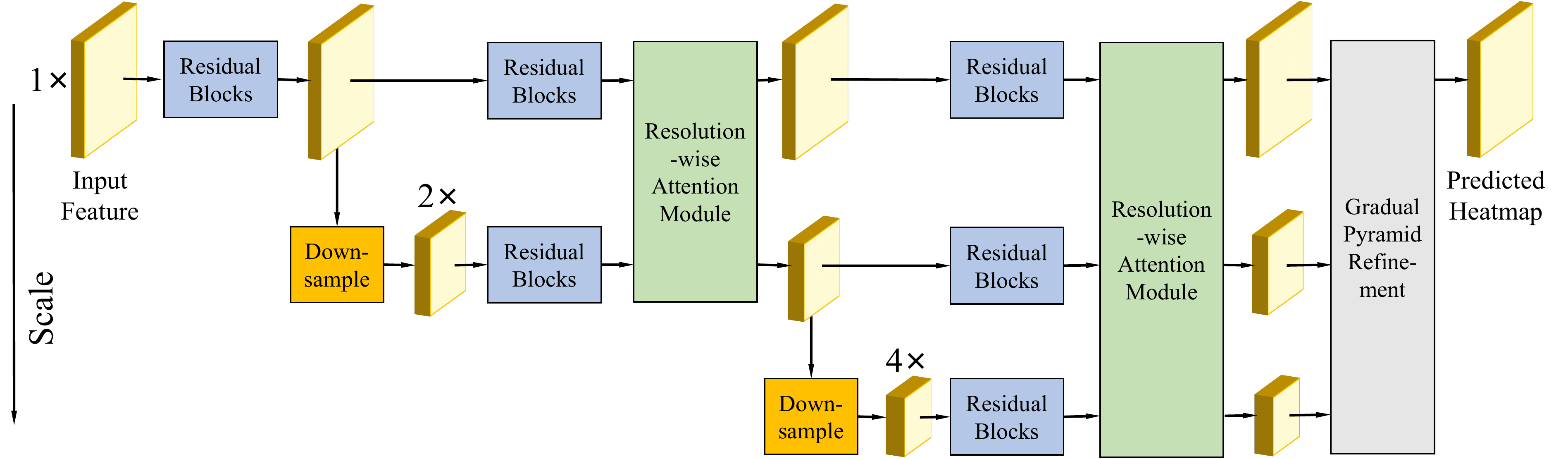}}
\caption{Architecture of our proposed network. The network starts from a subnetwork of high resolution at the first stage, and gradually adds low-resolution subnets in following stages. Resolution-wise Attention Module is applied for information exchange between features of different resolutions, and Gradual Pyramid Refinement module predicts human keypoint heatmap.} 
\label{whs}
\end{figure*}

As Fig.\ref{whs} shows, our whole network structure starts from a subnetwork of the highest resolution at the first stage, and gradually adds low-resolution subnets in following stages. The feature map of each subnetwork is extracted by residual blocks, after which the Resolution-wise Attention Modules integrates resolution-wise features for information exchange. Finally, the Gradual Pyramid Refinement fuses all features of all resolutions to predict human keypoint heatmap.

We empirically demonstrate the performance of proposed network on MS-COCO benchmark \cite{coco}. We have achieved state-of-the-art performance on COCO benchmark without using extra human keypoint training dataset.

\section{RELATED WORK}

{\noindent\bf Human Pose Estimation.}
Human pose estimation remains an active research topic for decades. Conventional approaches formulate this problem as graphic or tree models \cite{dgraph,tree08}, which use handcraft features for human keypoint predictions. The progress on pose estimation has been benefited by the strong capability of CNNs in the past few years. Recent powerful networks \cite{cpn1,hourglass,hrv1} represent joint positions with Gaussian peaks and estimate keypoint heatmaps for pixel-level joint locations.

{\noindent\bf Attention Mechanisms.}
Attention modules initially became popular on natural language processing tasks \cite{attentionall}. They have also been widely applied on image processing field, such as image classification \cite{hcgnet}, semantic segmentation \cite{danetseg}, and object detection \cite{tip19detect}. Different from previous methods that work on spatial or channel-wise feature maps \cite{senet, cbam} , we evaluate the importance of features accross different scales for more efficient resolution-wise information exchange.

{\noindent\bf Multi-Resolution Nets.}
State-of-the-art works on pixel-level tasks such as semantic segmentation \cite{refinenet}, crowd counting \cite{crowd2}, pose estimation \cite{hrv1}, and facial landmark detection \cite{robustface} usually combine feature maps from various resolutions for precise pixel labeling performance. There are two mainstreams of multi-resolution networks: one recovers high-resolution features from low-resolution representations \cite{hourglass}; the other maintains high resolution representations among the whole network architecture \cite{hrv1}. Our approach follows the second mainstream.

\section{Method}
\label{approach}

To enhance resolution-wise features efficiently, we propose two modules, Resolution-wise Attention Module (RAM) and Gradual Pyramid Refinement (GPR), to estimate the importance of features from different scales and combine resolution-wise information more efficiently for precise keypoint prediction. We will elaborate our proposed approaches in this section.
\subsection{Resolution-wise Attention Module}

The importance of features across resolutions differs under various contexts. Low-resolution feature maps are usually with large receptive fields and contribute to invisible human keypoint inference, while high-resolution features provide pixel-level details for accurate pose estimation. We propose Resolution-wise Attention Module to learn the importance of contextual features from multiple resolutions, which is applied for resolution-wise feature fusion. 

The inputs of a Resolution-wise Attention Module are feature maps of $M$ resolutions, i.e., ${\rm X=\{{X}_1,{X}_2,...,{X}}_M\rm\}$ and the output feature maps are ${\rm{Y}=\{{Y}_1,{Y}_2,...,{Y}}_N\rm\}$ of $N$ resolutions. For each output ${\rm Y}_i$, we firstly transform the input features ${\rm X=\{{X}_1,{X}_2,...,{X}}_M\rm\}$ to resolution $i$ with a set of sampling functions ${\rm T}_i = \{{\rm T}_1^i,{\rm T}_2^i,...,{\rm T}_M^i\}$, and then Resolution-wise Attention Module learns nonlinear scalars ${\rm{W}}^{i}=\{{W}_{1}^i,{W}_{2}^i,...,{W}_{M}^i\}$ as attention weights to represent the importance of different resolution-wise feature maps. 
\begin{figure}
\centerline{\includegraphics[width=8.5cm]{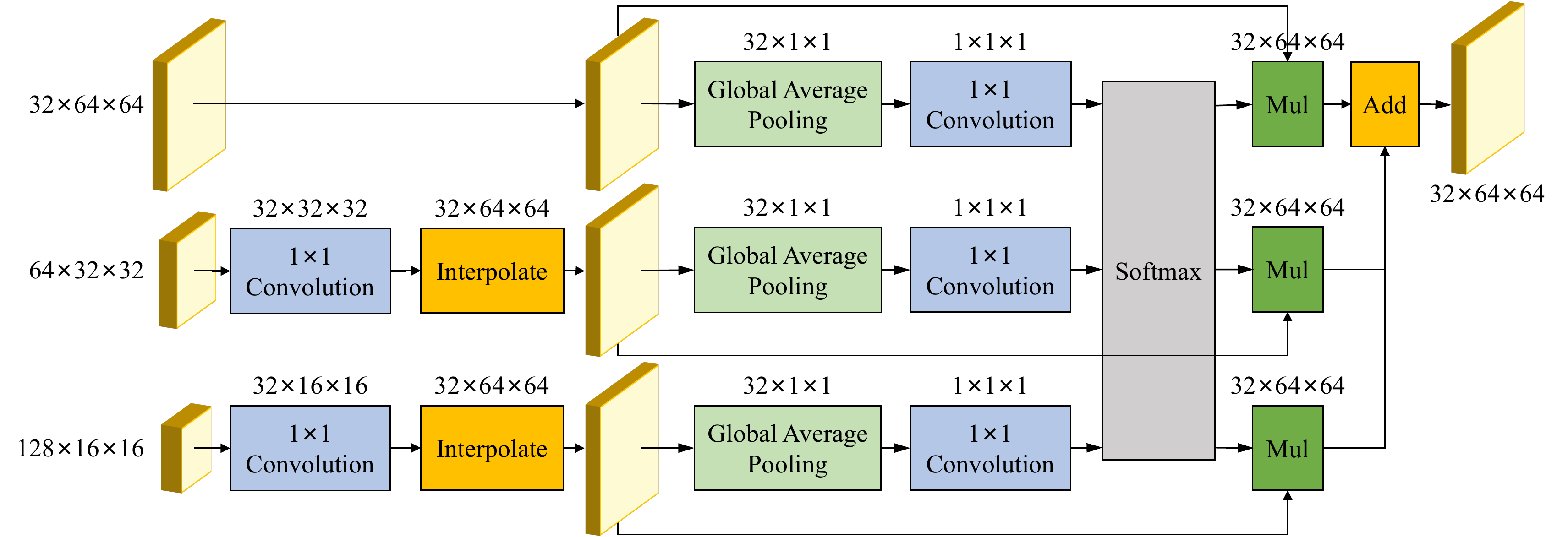}}
\caption{An example of the first branch in a Resolution-wise Attention Module with three inputs, in which feature maps of three different resolutions ${\rm X=\{{X}_1,{X}_2,{X}}_3\rm\}$ are aggregated to output ${\rm Y}_1$ of the same resolution.} \label{ratt}
\end{figure}

In detail, Resolution-wise Attention Module aggregates rescaled feature map ${\rm{X}}_h^i={\rm T}_h^i({\rm X}_h)$ with global pooling and $1\times1$ convolutions, i.e.,
\begin{equation}
E_h^i={\rm Conv^{1\times 1}}({\rm GlobalPool}({\rm{X}}_h^i)),
\end{equation}
where $E_h^i$ is a scalar with contextual information from the $h$th resolution, and $h\in\{1,2,...M\}$ denotes the index of $M$ input feature maps from different resolutions. The attention weights are then activated with SoftMax function to obtain nonlinearity, and we apply affine transformation to the activated scalas, so that the learnt weights are no longer limited in range of $(0, 1)$. The resolution-wise attention weights ${\rm{W}}^{i}=\{{W}_{1}^i,{W}_{2}^i,...,{W}_{M}^i\}$ are calculated by:
\begin{equation}
{W}_{h}^i=\frac{\exp({E_h^i})\cdot\omega_h^i}{\sum_{i=1}^{M}{E_h^i}}+\beta_h^i,
\end{equation}
where ${W}_{h}^i$ represents the importance of feature map from the $h$th resolution, and $\omega_h^i$ and $\beta_h^i$ are weight and bias for affine transformation. The output feature map ${\rm Y}_i$ of Resolution-wise Attention Module is the sum of multiplications of all rescaled feature maps and learned weights ${W}_{h}^i$, i.e.,
\begin{equation}
{\rm Y}_i=\sum_{i=1}^{M}{\rm T}_h^i({\rm{X}}_h)\cdot{W}_{h}^i.
\label{newfuse}
\end{equation}

Fig.\ref{ratt} displays an example of the first branch in Resolution-wise Attention Module, in which feature maps from three resolutions are aggregated to the first output resolution. 
\subsection{Gradual Pyramid Refinement}

The method of regressing human keypoint heatmaps counts for the performance of single person human pose estimation. In order to fuse the feature maps more efficiently at the end of network structure, we design Gradual Pyramid Refinement (GPR) to regresses heatmap by hierarchically merging feature maps from low to high resolutions, which is smilar to RefineNet \cite{refinenet}. 

\begin{figure}
\centerline{\includegraphics[width=7.2cm]{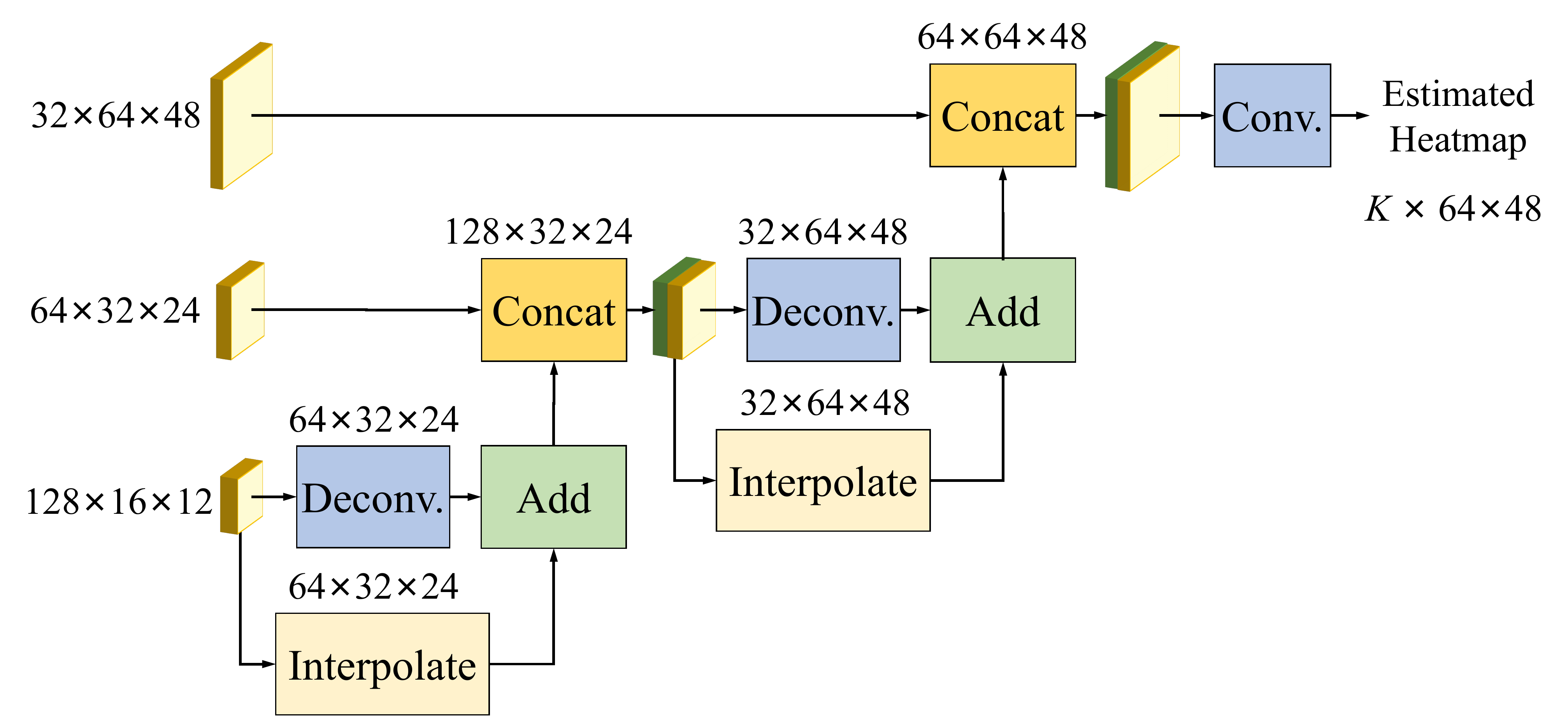}}
\caption{Architecture of a GPR module with three inputs, which regresses heatmaps by merging features of different levels across low to high resolution.} \label{gpr}
\end{figure}

The input of our Gradual Pyramid Refinement module are features maps ${\rm X=\{{X}_1,{X}_2,...,{X}}_M\rm\}$ of $M$ resolutions, which are progressively upsampled and combined from low to high resolutions. In our GPR module, each feature map from the $k$-th ($k>1$) resolution is upsampled by deconvolution and interpolation operations. Then the two upsampled feature maps are added together, whose adding result is concatenated with feature map of the $(k-1)$-th resolution. Those heatmaps are recursively integrated from low to high resolutions by the means of Equation \ref{fusesum} and Equation \ref{fusecat}, and we finally regress combined feature map of the first resolution with convolution operation to predict human keypoint heatmap $\rm{H}$, i.e.,
\begin{align}
{\rm{X}}_{k-1}'&={\rm Int}({\rm{X}}_k)+{\rm Deconv}({\rm{X}}_k),\label{fusesum}\\
{\rm{X}}_{k-1}&={\rm Concat}({\rm{X}}_{k-1}', {\rm{X}}_{k-1}),\label{fusecat}\\
{\rm{H}}\ \ \ \ \ &={\rm Conv}({\rm{X}}_{1}).\label{fusefinal}
\end{align}

Fig.\ref{gpr} demonstrates an example of GPR architecture, and we finally apply Mean Squared Error (MSE) as loss function for our keypoint heatmap regression. The performance of our our network will be demonstrated by experimental studies in Section \ref{sec:experiments}.

\section{Experiments}
\label{sec:experiments}
\subsection{Dataset and Metric}
We illustrate the effectiveness of our proposed methods on MS-COCO keypoint benchmark \cite{coco}. The COCO dataset contains more than 200k images and 250k person instances. Our networks are pretrained on ImageNet \cite{imgnet} dataset and finetuned on COCO train2017 dataset. We demonstrate the performance of our proposed network on COCO val2017 and test-dev2017 sets for comparisons with public state-of-the-art methods. We use OKS-based Mean Average Precision \cite{coco} (AP score) as evaluation metric on COCO dataset, where OKS calculates similarities between predicted keypoints and ground truth positions.

\begin{table*}
\begin{center}
{\caption{Comparing the performance of our network with other SOTA methods on COCO val2017 and test-dev2017 sets.}\label{sotacmp}}
\begin{tabular}{lccccccc}
\hline\rule{0pt}{12pt}
\rule{0pt}{12pt}Approach&Dataset&Input Size&Param.&FLOPs&CPU Time&GPU Time&AP\\
\hline
\\[-6pt]
\quad Stacked Hourglass \cite{hourglass}&val2017&$256\times192$&25.1M&14.3G&1470ms&11ms&66.9\\
\quad CPN + OHKM \cite{cpn1}&val2017&$256\times192$&27.0M&6.2G&-&-&69.4\\
\quad SimpleBaseline \cite{simplebaseline}&val2017&$256\times192$&68.6M&15.7G&490ms&7ms&72.0\\
\quad HRNet W32 \cite{hrv1}&val2017&$256\times192$&28.5M&7.1G&294ms&6ms&74.4\\	
\quad HRNet W48 \cite{hrv1}&val2017&$256\times192$&63.6M&14.6G&455ms&9ms&75.1\\
\quad \textbf{Ours of W32}&val2017&$256\times192$&31.4M&7.7G&345ms&7ms&\textbf{76.0}\\
\quad \textbf{Ours of W48}&val2017&$256\times192$&70.0M&15.8G&502ms&12ms&\textbf{76.5}\\
\hline
\quad SimpleBaseline \cite{simplebaseline}&val2017&$384\times288$&68.6M&35.6G&1053ms&11ms&74.3\\
\quad HRNet W32 \cite{hrv1}&val2017&$384\times288$&28.5M&16.0G&714ms&13ms&75.8\\
\quad HRNet W48 \cite{hrv1}&val2017&$384\times288$&63.6M&32.9G&1136ms&15ms&76.3\\
\quad \textbf{Ours of W32}&val2017&$384\times288$&31.4M&17.2G&877ms&17ms&\textbf{77.3}\\
\quad \textbf{Ours of W48}&val2017&$384\times288$&70.0M&35.6G&1352ms&23ms&\textbf{77.7}\\
\hline
\quad Ensembled CPN \cite{cpn1}&test-dev2017&$384\times288$&-&-&-&-&73.0\\
\quad SimpleBaseline \cite{simplebaseline}&test-dev2017&$384\times288$&68.6M&35.6G&1053ms&11ms&73.7\\
\quad HRNet W32 \cite{hrv1}&test-dev2017&$384\times288$&28.5M&16.0G&714ms&13ms&74.9\\
\quad HRNet W48 \cite{hrv1}&test-dev2017&$384\times288$&63.6M&32.9G&1136ms&15ms&75.5\\
\quad \textbf{Ours of W32}&test-dev2017&$384\times288$&31.4M&17.2G&877ms&17ms&\textbf{76.5}\\
\quad \textbf{Ours of W48}&test-dev2017&$384\times288$&70.0M&35.6G&1352ms&23ms&\textbf{77.0}\\
\hline
\\[-6pt]
\end{tabular}
\end{center}
\end{table*}
\subsection{Network Architecture}

The overall architecture of our whole network is shown in Fig.\ref{whs}. Our network architecture starts from a subnetwork of the highest resolution at the first stage, and other subnetworks of lower resolutions are gradually added in following stages. In each stage of our network, feature maps are extracted by residual blocks, then Resolution-wise Attention Modules are adopted for resolution-wise feature enhancements. The keypoint heatmap for pose estimation is finally predicted by Gradual Pyramid Refinement module. 

\subsection{Training and Testing}
For training setting, the human detection bounding box of our network is made to a fixed aspect ratio. i.e., $height:weight=4:3$. Our data augmentation operations include random scaling $([-35, 35])$, random rotation $([-45^\circ, 45^\circ])$, and flipping, which are the same as methods in \cite{simplebaseline} and \cite{hrv1}, and we adopt cutout \cite{cutout} for data augmentation. We employ SGDR \cite{sgdr} with initial learning rate of $0.001$, $T_0=16,$ and $T_{mul}=2$ as optimizer. Our models are trained for with batch size 32 on a computer with Intel Xeon E5 2620 V3 CPU and 4 Nvidia Titan X GPUs.

For validation, we use a two-stage paradigm on COCO benchmark: we firstly detect persons and then estimate keypoints of detected human bodies. We use Faster-RCNN \cite{fastercnn} human detector, which is the same as the detector of \cite{simplebaseline} and \cite{hrv1} for fair comparison. Following common practice, the keypoints are predicted by averaging keypoint heatmaps of original and flipped images. A quarter offset from the highest to the second highest response is used to predict final keypoints. Inference time for each person is tested on the same platform as mentioned above. We didn't report the inference time of CPN as we didn't implement it on our platform.

\subsection{Component Ablation Studies}
We use our network with channels of 32 as backbone with input of $256\times 192$ in ablation studies, the effectiveness of our modules is shown on COCO val2017 dataset. The AP score decreases from 76.0 to 75.7 if we remove the attention weights of RAMs. Apart from that when we replace GPR module with directly rescaling and summing operations, the AP score decreases from 76.0 to 75.2. The cutout operation helps the increment of AP score by 0.6. 

\subsection{Comparisons with SOTA Approaches}
We compare the performance of our network with other state-of-the-art approaches on COCO val2017 and test-dev sets, and we demonstrate performance on Table \ref{sotacmp}. Our small network with 32 channels performs even better than HRNet W48 under same resolutions with nearly half of its parameters and FLOPs. Our largest model with input size $384\times 288$ achieves 77.7 AP score on COCO val2017 set and 77.0 on COCO test-dev 2017 set without training on extra human keypoint datasets.

\section{Conclusion}
In this paper, we propose to use Resolution-wise Attention Module and Gradual Pyramid Refinement to learn enhanced resolution-wise feature maps for precise pose estimation. Resolution-wise Attention Module learns a group of weights to represent the importance of features across resolutions, and the Gradual Pyramid Refinement module gradually merges every two feature maps from low to high resolutions to regress more accurate human keypoint heatmaps. We have achieved state-of-the-art performance at 77.7 AP score on COCO val2017 dataset and 77.0 AP score on COCO test-dev2017 dataset without using any extra keypoint training data. 

\vfill
\pagebreak

\bibliographystyle{IEEEbib}
\bibliography{refs}

\end{document}